\documentclass[fleqn,11pt]{wlscirep}
\usepackage[utf8]{inputenc}
\usepackage[T1]{fontenc}
\usepackage{bm}
\usepackage{xspace}
\usepackage{xcolor}
\usepackage{amsmath,amssymb}
\usepackage{mathtools}

\usepackage{float}
\usepackage{siunitx}
\DeclareSIUnit{\microsiemens}{\micro \siemens}
\DeclareSIUnit{\nJ}{\nano \joule}
\usepackage{multirow}
\usepackage{algorithm}
\usepackage{algpseudocode}
\usepackage{nccmath}
\usepackage[scr=rsfso,cal=cm]{mathalfa}

\usepackage[font=small, justification=justified]{caption}

\algnewcommand{\LineComment}[1]{\Statex \hskip\algorithmicindent \(\triangleright\) #1}

\usepackage{amsthm}  

\usepackage{xr}
\makeatletter

\newcommand*{\addFileDependency}[1]{
\typeout{(#1)}
\@addtofilelist{#1}
\IfFileExists{#1}{}{\typeout{No file #1.}}
}\makeatother

\newcommand*{\myexternaldocument}[1]{%
\externaldocument{#1}%
\addFileDependency{#1.tex}%
\addFileDependency{#1.aux}%
}
\myexternaldocument{supplementary}

\title{Driving risk emerges from the required two-dimensional joint evasive acceleration}

\author[1]{Hao Cheng}
\author[1]{Yanbo Jiang}
\author[2]{Wenhao Yu}
\author[3]{Rui Zhou}
\author[3]{Jiang Bian}
\author[1]{Keyu Chen}
\author[1]{Zhiyuan Liu}
\author[4]{Heye Huang}
\author[5]{Hailun Zhang}
\author[2]{Fang Zhang}
\author[1]{Jianqiang Wang}
\author[1,*]{Sifa Zheng}

\affil[1]{School of Vehicle and Mobility, Tsinghua University, Beijing 100084, China}
\affil[2]{State Key Laboratory of Intelligent Green Vehicle and Mobility, Beijing 100084, China}
\affil[3]{School of Traffic \& Transportation Engineering, Central South University, Changsha 410000, China}
\affil[4]{Singapore-MIT Alliance for Research and Technology (SMART), Massachusetts Institute of Technology, Singapore 138602, Singapore}
\affil[5]{School of Automotive Engineering, Chang'an University, Xi'an 710064, China}

\begin{abstract}
\fontsize{10.5pt}{12.5pt}\selectfont
Most autonomous-driving safety benchmarks use time-to-collision (TTC) to assess risk and guide safe behaviour. However, TTC-based methods treat risk as a one-dimensional closing problem, despite the inherently two-dimensional nature of collision avoidance, and therefore cannot faithfully capture risk or its evolution over time. Here, we report evasive acceleration (EA), a hyperparameter-free and physically interpretable two-dimensional paradigm for risk quantification. By evaluating all possible directions of collision avoidance, EA defines risk as the minimum magnitude of a constant relative acceleration vector required to alter the relative motion and make the interaction collision-free. Using interaction data from five open datasets and more than 600 real crashes, we derive percentile-based warning thresholds and show that EA provides the earliest statistically significant warning across all thresholds. Moreover, EA provides the best discrimination of eventual collision outcomes and improves information retention by 54.2--241.4\% over all compared baselines. Adding EA to existing methods yields 17.5--95.5 times more information gain than adding existing methods to EA, indicating that EA captures much of the outcome-relevant information in existing methods while contributing substantial additional nonredundant information.
Overall, EA better captures the structure of collision risk and provides a foundation for next-generation autonomous-driving systems.
\end{abstract}

\begin{document}
\setcounter{secnumdepth}{0}
\flushbottom
\maketitle

\begin{figure}[H]
    \centering
    \includegraphics[width=0.95\textwidth]{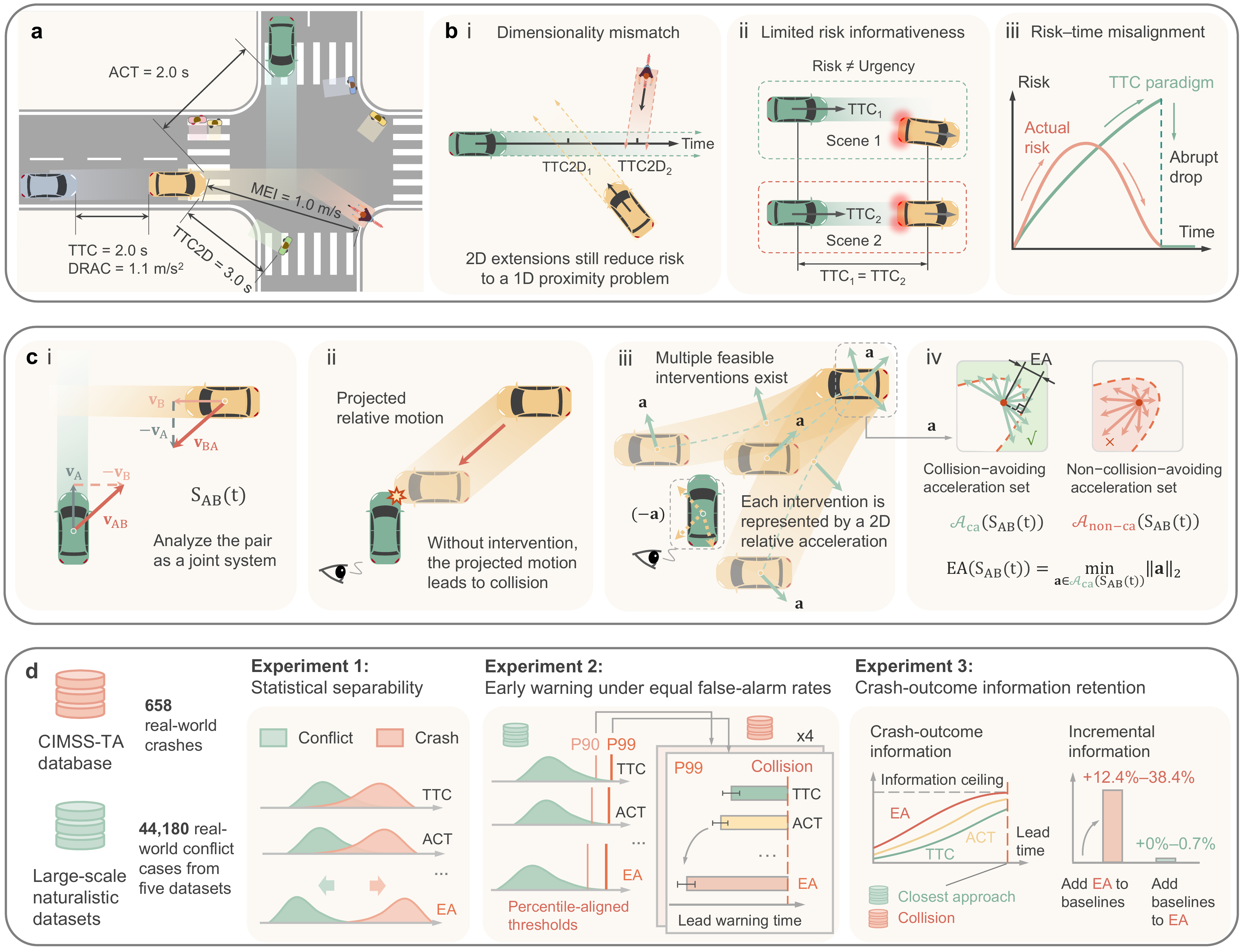}
    \caption{
    \textbf{Evasive acceleration (EA) as a new paradigm for risk quantification: motivation, concept and validation.}
    \textbf{a,} Illustration of a traffic scene evaluated by TTC, its variants, and existing risk quantification methods. TTC and its derivatives have long shaped safety assessment across training, testing, and regulatory settings.
    \textbf{b,} Conceptual illustration of the core limitation of the TTC paradigm. i, TTC evaluates risk urgency solely through a single temporal dimension, creating a dimensionality mismatch. ii, As a result, scenes with the same TTC value can differ substantially in actual risk, indicating limited risk informativeness. iii, TTC can also become misaligned with the temporal evolution of risk; for example, it may peak as a conflict resolves and then drop abruptly, yielding a distorted safety gradient.
    \textbf{c,} Conceptual illustration of EA. i, Example scenario. ii, In relative motion, the yellow vehicle approaches the green vehicle, and without intervention the interaction would lead to a collision. iii, We model this intervention as a constant relative acceleration that alters the relative motion of the two vehicles to ensure collision avoidance. Relative accelerations of different magnitudes and directions can resolve the conflict. For visualisation, the relative acceleration is shown as acting solely on the yellow vehicle, whereas it is in fact defined for the joint system of both vehicles. iv, Valid interventions form a collision-avoiding set, and the EA value is defined as the minimum magnitude among all relative acceleration vectors in this set.
    \textbf{d,} Validation framework. Using 658 reconstructed real crashes and 44,180 potential-conflict events from five naturalistic trajectory datasets, we conduct three complementary experiments to assess statistical separability, early-warning timeliness, and retained information about crash outcomes. Across all three experiments, EA outperforms existing baselines.
    }
    \label{fig:Introduction_main}
\end{figure}

Autonomous driving is among the most promising technologies of the twenty-first century~\cite{feng2026breaking}. However, improving safety remains a central and enduring challenge~\cite{kolekar2020human,feng2023dense,qian2026test}. Across both academia and industry, time-to-collision (TTC) has become the dominant method for quantifying driving risk. Most existing testing platforms and benchmarks, including nuPlan~\cite{dauner2023parting} and NAVSIM~\cite{dauner2024navsim,cao2025pseudo}, rely on TTC to evaluate safety and guide autonomous driving systems~\cite{liu2025behavioral,cui2025vp}. As a result, TTC broadly shapes data curation and generation~\cite{fan2024risk,guo2025ipad} as well as model training~\cite{li2024hydra,yao2025drivesuprim,guo2025ipad,li2025generalized,song2025diver,reddy2026rapid,yangread,li2026plannerrft,song2026senna}. In recent years, regulations and technical standards in several countries have also incorporated TTC into formal safety frameworks~\cite{nolte2025review,de2024scenario,olleja2025validation,rasch2020drivers,nodine2019indicators,singh2025scenario} to assess and constrain the performance of these systems. In this sense, TTC is not merely one risk quantification method among many, but the paradigm that has fundamentally shaped how practitioners understand driving risk and has steered the direction of safety development over the past decade (Fig.~\ref{fig:Introduction_main}a).

However, the TTC paradigm is fundamentally flawed. For scenes that differ substantially in evasive difficulty and underlying risk, TTC can assign nearly identical values, a phenomenon we term limited risk informativeness (Fig.~\ref{fig:Introduction_main}b ii). TTC can also misrepresent the temporal evolution of risk. Specifically, it may peak precisely at the instant just before a conflict resolves and then jump discontinuously to infinity, implying an abrupt drop from peak risk to zero. We term this phenomenon risk--time misalignment (Fig.~\ref{fig:Introduction_main}b iii). We argue that both phenomena arise from the deeper methodological problem of dimensionality mismatch, where an essentially one-dimensional risk quantification paradigm is used to represent two-dimensional, multidirectional interaction risk (Fig.~\ref{fig:Introduction_main}b i). Recent studies have likewise identified a range of limitations and practical failure modes of TTC. Some have proposed localised modifications or derivative variants of TTC~\cite{venthuruthiyil2022anticipated,guo2023modeling,fu2024research,wang2024surrogate,jafari2024pedestrians,behera2025improved,lin2025predicted,tang2025analysis}. Although these methods can improve performance in specific scenarios, they fail to resolve the core problem of quantifying two-dimensional risk with a one-dimensional paradigm. Others have shown that TTC can be problematic in modern autonomous driving safety applications. Specifically, TTC-based warning systems may generate large numbers of false alarms, thresholds intended to identify high-risk events may capture many safe events~\cite{yan2024evaluation,choi2025predictive,biswas2024modeling}, and TTC may fail to provide an appropriate safety gradient for training or evaluation~\cite{wu2025ai2,lu2025risk,diwakar2023evaluating,do2025evaluation,elsamadisy2024safe,westhofen2023criticality}.

Addressing the structural limitations of the TTC paradigm requires a shift towards a more faithful quantification of multidirectional interaction risk. We propose evasive acceleration (EA), a two-dimensional risk quantification paradigm that directly measures the minimum instantaneous cost of collision avoidance (Fig.~\ref{fig:Introduction_main}c). Unlike prevailing approaches that quantify risk along a predefined direction, EA evaluates all possible directions of relative collision avoidance and retains the least costly one. Consider the scenario in Fig.~\ref{fig:Introduction_main}c i. From the perspective of the green vehicle, the yellow vehicle approaches (Fig.~\ref{fig:Introduction_main}c ii), and without intervention their interaction would result in a collision. To avert this outcome, we introduce a hypothetical two-dimensional constant relative acceleration of arbitrary magnitude and direction that directly alters the relative motion of the joint system (Fig.~\ref{fig:Introduction_main}c iii). At any non-collision instant, the minimum magnitude among all collision-avoiding relative acceleration vectors defines the EA value (Fig.~\ref{fig:Introduction_main}c iv). Notably, EA is defined at the level of relative motion, rather than for either road user individually. Dynamically, it captures the continuous evolution of risk, from its escalation before an evasive manoeuvre to its dissipation once the manoeuvre takes effect. EA is also computationally transparent, hyperparameter-free, and physically intuitive. We further develop an efficient computational framework that achieves an average single-frame computation time of 5~ms, enabling direct large-scale deployment.

We assemble a large and heterogeneous evaluation set (Fig.~\ref{fig:Introduction_main}d) comprising 44,180 naturalistic interactions from five open datasets across Germany, China, and the United States, alongside 658 reconstructed real-world crashes from the CIMSS-TA database~\cite{zhou2024would,zhou2025crash}. The naturalistic data cover a wide range of scenarios, including highways and urban intersections, and involve various road users, such as passenger cars, trucks, cyclists, and pedestrians, whereas the crash data provide rare trajectory-level ground truth for high-risk scenarios. On this basis, we conduct three rounds of validation, each targeting a core capability required for a risk quantification method.

First, we evaluate the ability of a risk quantification method to statistically distinguish crash precursors from routine noncrash interactions (Fig.~\ref{fig:Introduction_main}d, Experiment~1). For noncrash interactions, we use the peak value of each risk quantification method within each event. For crash data, we define four precrash lead windows. Across all four windows, EA consistently achieves the best AUPRC and AUROC among the compared methods. For example, in the representative precrash window of $[-1.5, -0.1]$~s, EA attains an AUPRC of 0.901 and an AUROC of 0.934, whereas the compared baselines range from 0.204 to 0.812 in AUPRC and from 0.458 to 0.865 in AUROC.

Second, we investigate early warning timeliness in a practical setting, determining which risk quantification method provides the earliest and most sustained warning under a fixed false-alarm budget without relying on learning-based prediction modules (Fig.~\ref{fig:Introduction_main}d, Experiment~2). To ensure fairness, we calibrate warning thresholds from the naturalistic interaction data. Specifically, we use the 90th, 95th, 99th, and 99.5th percentiles of the output distribution of each risk quantification method as warning thresholds. Replaying the crash data reveals that EA provides statistically earlier warnings across all thresholds. Moreover, its relative advantage increases as the false-alarm constraint becomes stricter. At the 99.5th percentile threshold, EA provides warnings 120--267\% earlier than TTC-based methods. This superior warning performance stems directly from EA's ability to distinguish benign routine approaches from precollision hazards.

Third, we assess how much information about the eventual crash-versus-noncrash outcome is retained in each method's risk value (Fig.~\ref{fig:Introduction_main}d, Experiment~3). Specifically, using 658 crashes and 40,417 time-aligned noncollision potential conflicts, we quantify how much uncertainty about the eventual outcome is reduced by each method's risk value for the same instantaneous driving scene. Under a cross-entropy framework, EA retains the most collision-outcome-relevant information across the preevent horizon and yields the lowest residual uncertainty about the eventual outcome. Compared with both classical and recent advanced baselines, EA improves information retention by 54.2--241.4\%. Moreover, EA exhibits strong directional asymmetry in incremental information, with asymmetry ratios of 17.5--95.5$\times$. Adding EA to existing risk quantification methods contributes an additional 12.4--38.4\% of the information ceiling, whereas adding other methods to EA yields negligible additional information. This indicates that EA captures nearly all collision-outcome-relevant information contained in the existing methods while contributing substantial nonredundant information beyond them.

Taken together, these results show that the EA paradigm more closely reflects the underlying nature of collision risk, providing a physically intuitive safety foundation for next-generation autonomous driving systems.

\section{Results}
\subsection{Evasive acceleration quantifies the minimum intervention required to avoid collision}
\label{subsec:ea_concept}

A traffic conflict refers to an interaction in which extrapolating the current state of motion would lead to a collision.~\cite{zheng2014traffic,johnsson2018search,sarkar2024review} We next formalise EA in relative motion space as the minimum constant relative acceleration required to keep the future interaction collision-free.

Consider two road users $A$ and $B$ at time $t$. Over a future interval of interest $\tau\in[t,t+T]$, let $\mathbf p_A^{0}(\tau)$ and $\mathbf p_B^{0}(\tau)$ denote the trajectories obtained by extrapolating their current motion without any additional evasive response. The relative trajectory is
\begin{equation}
\mathbf r^{0}(\tau):=\mathbf p_A^{0}(\tau)-\mathbf p_B^{0}(\tau),
\qquad \tau\in[t,t+T].
\label{eq:r0_results}
\end{equation}
Let $\mathcal C\subset\mathbb{R}^{2}$ denote the collision set in relative position space, that is, the set of relative configurations for which the occupied regions of the two road users overlap.

We then introduce a hypothetical constant relative evasive acceleration $\mathbf a\in\mathbb{R}^{2}$. Under this intervention, the relative trajectory becomes $\mathbf r^{0}(\tau)+\frac{1}{2}\mathbf a(\tau-t)^2$. EA is defined as the minimum magnitude of such intervention that keeps the future relative trajectory collision-free:
\begin{equation}
\mathrm{EA}\!\left(S_{AB}(t)\right)
=
\min_{\mathbf a\in\mathbb{R}^{2}}
\|\mathbf a\|_2
\quad
\mathrm{s.t.}
\quad
\mathbf r^{0}(\tau)+\frac{1}{2}\mathbf a(\tau-t)^2 \notin \mathcal C,
\ \forall \tau\in[t,t+T].
\label{eq:EA_results}
\end{equation}

By construction, EA quantifies the minimum constant relative acceleration required to keep the future relative trajectory collision-free. An EA of zero indicates that no additional intervention is required for collision avoidance, whereas larger EA values indicate that stronger evasive action is needed. We further develop an efficient computational framework for EA, as described in the \hyperref[subsec:ea_framework_methods]{Methods} section.

\begin{figure}[H]
    \centering
    \includegraphics[width=1.0\linewidth]{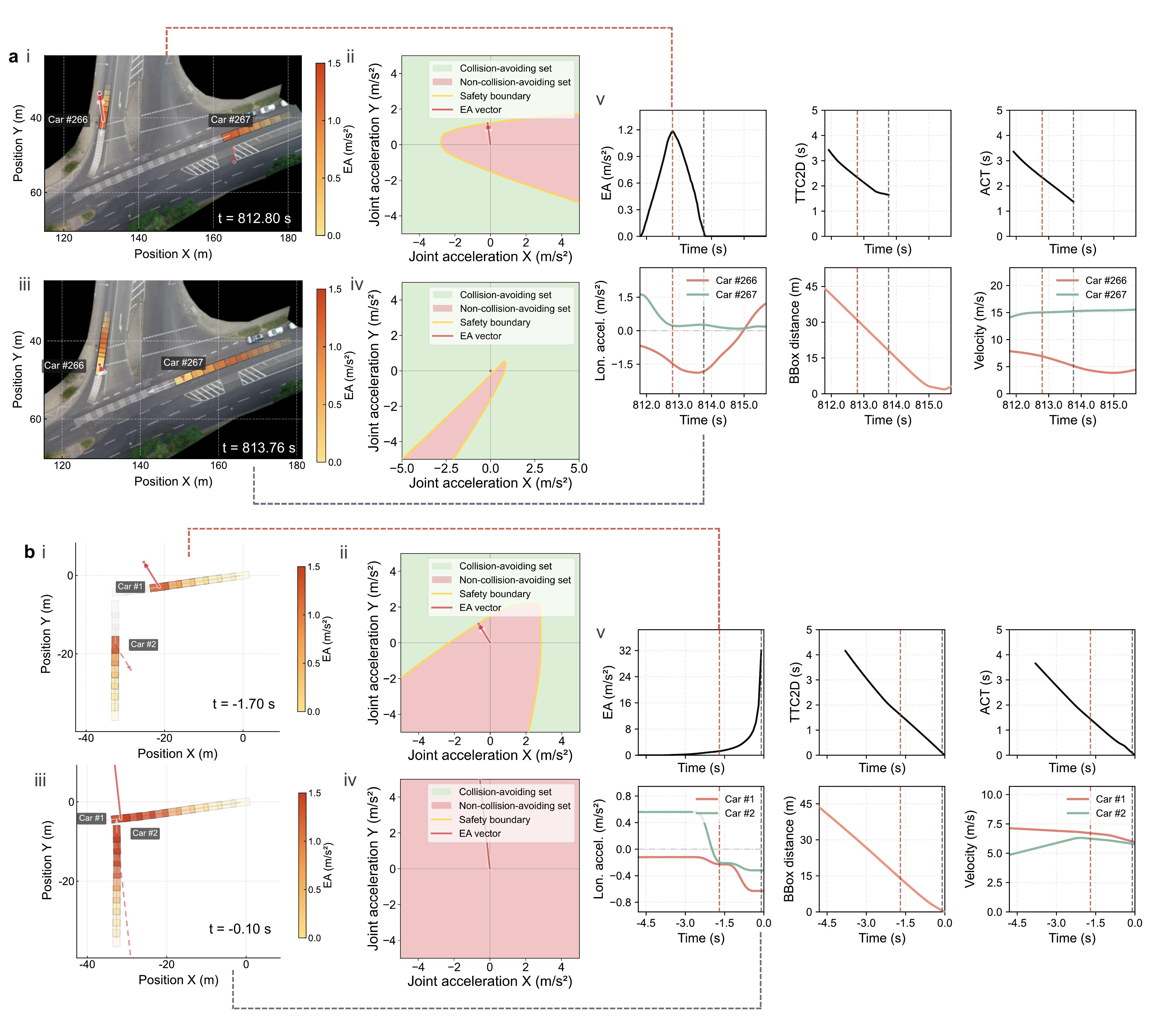}
    \caption{
        \textbf{Representative interaction cases showing how EA captures risk evolution and conflict resolution.}
        \textbf{a,} Merging interaction at an intersection. Panels i and iii show two representative instants of the interaction, with trajectory colours indicating EA values. Panels ii and iv show the corresponding collision-avoiding and non-collision-avoiding sets in two-dimensional joint-acceleration space. The red solid arrow denotes the EA vector, and its length represents the EA value. For visualisation, it is shown on one vehicle, whereas the red dashed arrow denotes its equal-and-opposite counterpart on the other vehicle. EA is independent of which vehicle is taken to execute the intervention or how the evasive effort is distributed between the two vehicles. Panel v shows the time series of EA, TTC2D, ACT, longitudinal acceleration, bounding-box distance, and velocity. The interaction is resolved when car \#266 yields by decelerating, and EA correspondingly first increases and then decreases.
        \textbf{b,} T-bone crash at an intersection in Hunan, China. EA increases continuously throughout the event. Panels i and iii show two representative precrash instants, and panels ii and iv show the corresponding collision-avoiding sets in two-dimensional joint-acceleration space. Panel v shows the time series of EA, TTC2D, ACT, longitudinal acceleration, bounding-box distance, and velocity.
        }
        \label{fig:Results_case}
\end{figure}

We next examine two representative cases to illustrate how EA quantifies risk in traffic interactions. In the merging interaction at the intersection shown in Fig.~\ref{fig:Results_case}a, car \#266 and car \#267 enter a conflict that is eventually resolved. As shown in Fig.~\ref{fig:Results_case}a i--ii, EA first increases to a peak of $1.21~\mathrm{m/s^2}$ and then decreases, capturing both the build-up of risk before an effective evasive response is established and the subsequent dissipation of risk once that response takes effect. Notably, when the longitudinal deceleration of car \#266 first reaches $-1.42~\mathrm{m/s^2}$, EA begins to decline, indicating that the evasive manoeuvre is now sufficient to reduce the interaction risk. In contrast, TTC-based methods do not effectively capture the extent to which the conflict has been resolved. The two TTC extensions, two-dimensional TTC (TTC2D) and anticipated collision time (ACT), both continue to decrease as the interaction evolves, suggesting progressively higher risk (see Fig.~\ref{fig:Results_case}a v). Once a collision is no longer detected, however, TTC-based methods abruptly jump to positive infinity, indicating the absence of collision risk. This pattern exemplifies the risk--time misalignment phenomenon described above. Specifically, Fig.~\ref{fig:Results_case}a iii--iv shows the instant of highest TTC-based risk, with $\mathrm{ACT}=1.37~\mathrm{s}$ and $\mathrm{TTC2D}=1.64~\mathrm{s}$, whereas EA is only $0.07~\mathrm{m/s^2}$.

Figure~\ref{fig:Results_case}b shows a T-bone crash that occurs in Hunan, China. At $1.70~\mathrm{s}$ before the collision, EA has already reached $1.25~\mathrm{m/s^2}$ and continues to increase throughout the interaction (see Fig.~\ref{fig:Results_case}b i--ii). By $0.1~\mathrm{s}$ before impact, EA has risen to $31.97~\mathrm{m/s^2}$, indicating that the collision has become unavoidable (see Fig.~\ref{fig:Results_case}b iii--iv). Over the course of the interaction, TTC-based methods exhibit an approximately linear decline, whereas EA shows a sharply accelerating nonlinear rise (see Fig.~\ref{fig:Results_case}b v). This case illustrates a general property of high-risk interactions: as risk increases, even the minimum effort required across all possible conflict-resolving directions becomes progressively greater.

Taken together, these two cases show that the specific instants shown in Fig.~\ref{fig:Results_case}a i--ii and Fig.~\ref{fig:Results_case}b i--ii correspond to a similar instantaneous risk level. The key difference is that the conflict in Fig.~\ref{fig:Results_case}a is subsequently resolved by timely and effective evasive action, whereas the conflict in Fig.~\ref{fig:Results_case}b continues to intensify in the absence of such action. This suggests that EA reflects the remaining intervention required to avoid collision. In addition, Supplementary Section~3a shows that TTC-based methods may overstate the risk of benign interactions that require only minimal evasive effort, whereas EA quantifies such cases more faithfully. We next evaluate whether this effort-based quantity provides superior crash discrimination, earlier warnings, and greater retention of collision-relevant information than existing methods.

\subsection{Statistical separability of crashes and potential conflicts}
\label{Statistical separability of crashes and potential conflicts}

To support statistically meaningful evaluation across heterogeneous interaction regimes, we assemble a large-scale potential conflict dataset from five open-source, high-resolution naturalistic driving datasets collected in Germany, China, and the United States. The datasets span highways and urban intersections and cover a diverse range of road users, including passenger cars, trucks, cyclists, motorcyclists, and pedestrians. Using deliberately conservative conflict screening thresholds selected within the range of existing approaches to ensure broad coverage of potential conflicts (as detailed in the \hyperref[potential_conflicts]{Methods} section), we extract 44,180 potential conflict cases: 3,376 from highD~\cite{krajewski2018highd}, 16,293 from exiD~\cite{moers2022exid}, 8,844 from inD~\cite{bock2020ind}, 10,386 from SIND~\cite{xu2022drone}, and 4,481 from the Lateral Conflict Resolution dataset based on Argoverse 2 (Argo2-LCR)~\cite{li2024lateral}. As positive samples, we use an independent set of 658 reconstructed passenger-vehicle crashes from the CIMSS-TA database~\cite{zhou2024would,zhou2025crash}.

A risk quantification method must, at minimum, assign higher risk to crash precursors than to routine noncrash interactions. However, evaluating only the final instant before collision is of limited value because near-contact geometry alone already makes classification easy. We therefore assess statistical separability under four progressively earlier lead windows to test whether crash precursors remain distinguishable before collision becomes visually or geometrically obvious.

We evaluate EA against a structured set of baseline methods representing the principal existing approaches to instantaneous risk quantification. These include the classical one-dimensional methods TTC and deceleration rate to avoid crash (DRAC)~\cite{almqvist1991use}, established two-dimensional extensions such as TTC2D~\cite{jiao2023ttc} and anticipated collision time (ACT)~\cite{venthuruthiyil2022anticipated}, a two-dimensional counterpart of DRAC referred to here as DRAC2D~\cite{huang2025estimating}, and the recently proposed modified emergency index (MEI)~\cite{cheng2025modified}. For certain analyses, we also include bounding-box distance as a simple spatial-proximity baseline. Because this study focuses on instantaneous risk quantification, post-encroachment time (PET)~\cite{allen1978analysis} is not included in the main comparison. We also evaluate time advantage (TAdv)~\cite{laureshyn2010evaluation}, but in crossing conflicts it tends to collapse to near-zero values even in low-risk scenarios, limiting its utility for statistical discrimination. Given that its primary value lies in illustrating case-specific risk evolution~\cite{cheng2025emergency,jiao2026learning}, TAdv results are not shown in the main text.

For each method, each potential conflict case contributes a single negative sample, defined as the maximum risk value attained within that case. This avoids temporal redundancy while ensuring that each noncrash interaction is represented once. Positive samples are taken from four precrash windows in the crash data, namely, \([-0.5, -0.1]\), \([-1.0, -0.1]\), \([-1.5, -0.1]\), and \([-2.0, -0.1]\)~s. Because the crash trajectories are sampled at 10~Hz, each crash contributes up to 5, 10, 15, and 20 positive frames under these four settings, respectively. This design does not require every positive sample to receive a higher risk value than every negative sample. Instead, it tests whether crash precursors are statistically shifted towards higher risk than routine interactions at the population level. Performance is evaluated using the following complementary criteria: area under the precision--recall curve (AUPRC), area under the receiver operating characteristic curve (AUROC), the Kolmogorov--Smirnov (KS) statistic, and recall (equivalently, true positive rate, TPR) at fixed false positive rate (FPR) operating points of 1\%, 5\%, and 10\%.

\begin{figure}[H]
    \centering
    \includegraphics[width=0.90\linewidth]{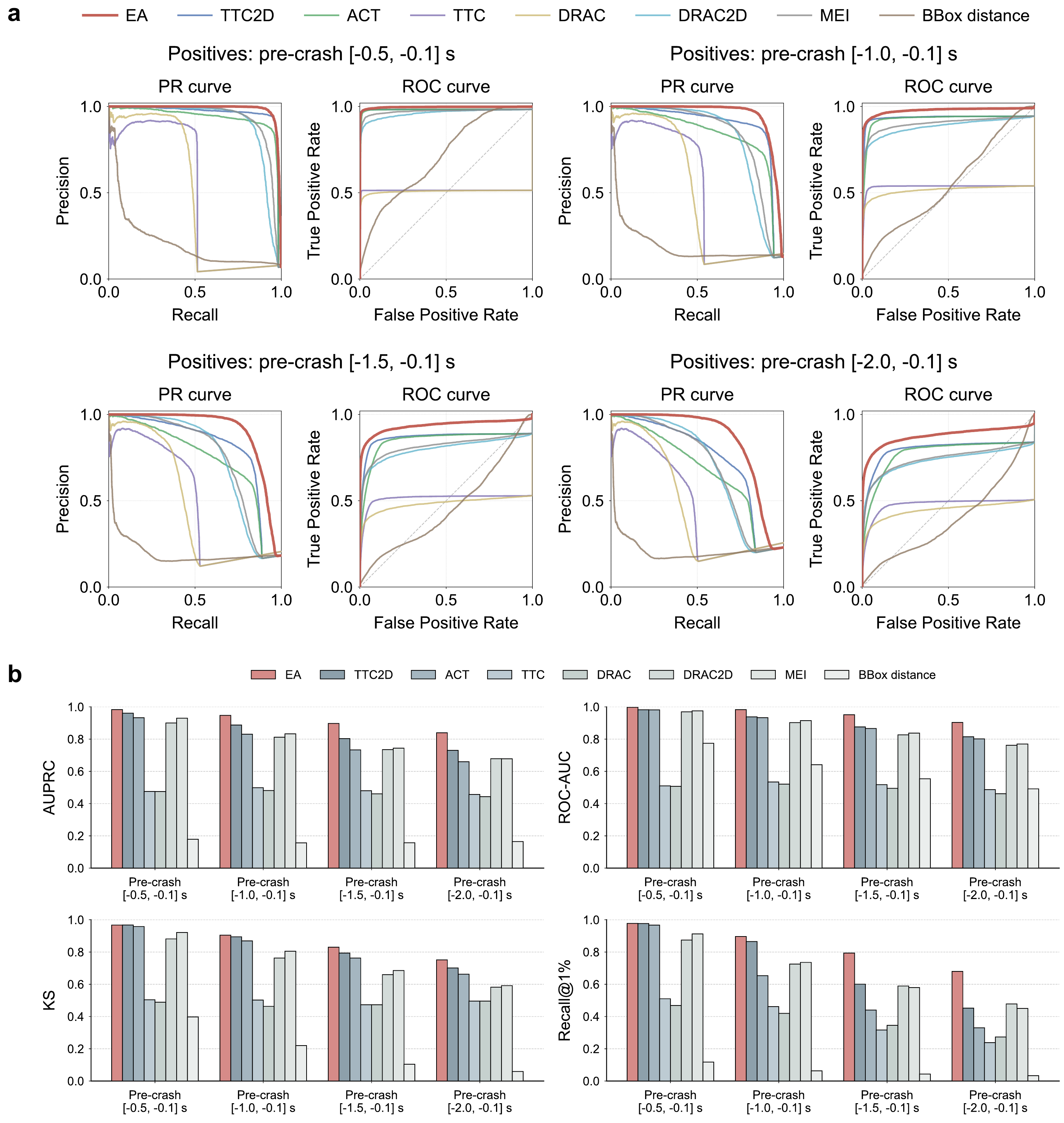}
    \caption{
        \textbf{Statistical separability of precrash states and potential conflicts.}
        \textbf{a,} Precision--recall (PR) and receiver operating characteristic (ROC) curves for four precrash lead windows: \([-0.5, -0.1]\), \([-1.0, -0.1]\), \([-1.5, -0.1]\), and \([-2.0, -0.1]\)~s. EA shows the strongest overall discrimination across the four lead windows, with its advantage becoming more pronounced at earlier precrash horizons.
        \textbf{b,} Summary discrimination metrics across the same four precrash lead windows, including AUPRC, AUROC, the KS statistic, and Recall@1\% FPR. Across these criteria, EA remains the leading overall performer, with its advantage becoming more pronounced at earlier precrash horizons.
    }
    \label{fig:Results_1_1}
\end{figure}

Figure~\ref{fig:Results_1_1} summarises the separability results across the four lead windows. Across both the PR/ROC curves and the summary discrimination metrics, EA shows the strongest overall performance among the compared methods. It achieves the best AUPRC and AUROC at every horizon and remains the leading overall performer across the reported criteria, with its advantage becoming more pronounced at earlier and more challenging precrash horizons. For example, in the representative precrash window of \([-1.5, -0.1]\)~s, EA attains an AUPRC of 0.901 and an AUROC of 0.934, whereas the compared baselines range from 0.204 to 0.812 in AUPRC and from 0.458 to 0.865 in AUROC. This advantage is also reflected under strict false-alarm constraints: at 1\% FPR, EA reaches a TPR of 0.759 in the same window, compared with 0.482 for TTC2D and 0.527 for DRAC2D. As the lead window moves further from impact, the relative advantage of EA becomes even clearer. In the earliest lead window, \([-2.0, -0.1]\)~s, EA still achieves an AUPRC of 0.848 and an AUROC of 0.875, compared with the second-best values of 0.746 and 0.795, both attained by TTC2D. Full results for all compared methods and lead windows are provided in Supplementary Table~3.

Taken together, these results show that EA provides the strongest threshold-independent statistical separability between precrash states and large-scale naturalistic potential conflicts. This is consistent with EA capturing aspects of safety criticality that are not reducible to geometric imminence alone.

\subsection{Early-warning timeliness under percentile-aligned false-alarm constraints}
\label{sec:results_wlt}
To evaluate early warning timeliness under realistic false-alarm constraints, we conduct a threshold-aligned early warning experiment in which each risk quantification method is used directly as a warning signal, without the aid of any black-box prediction module. It is not intended to benchmark a deployable warning system but to isolate the warning timeliness of the risk quantification method itself.

\begin{figure}[H]
    \centering
    \includegraphics[width=0.85\linewidth]{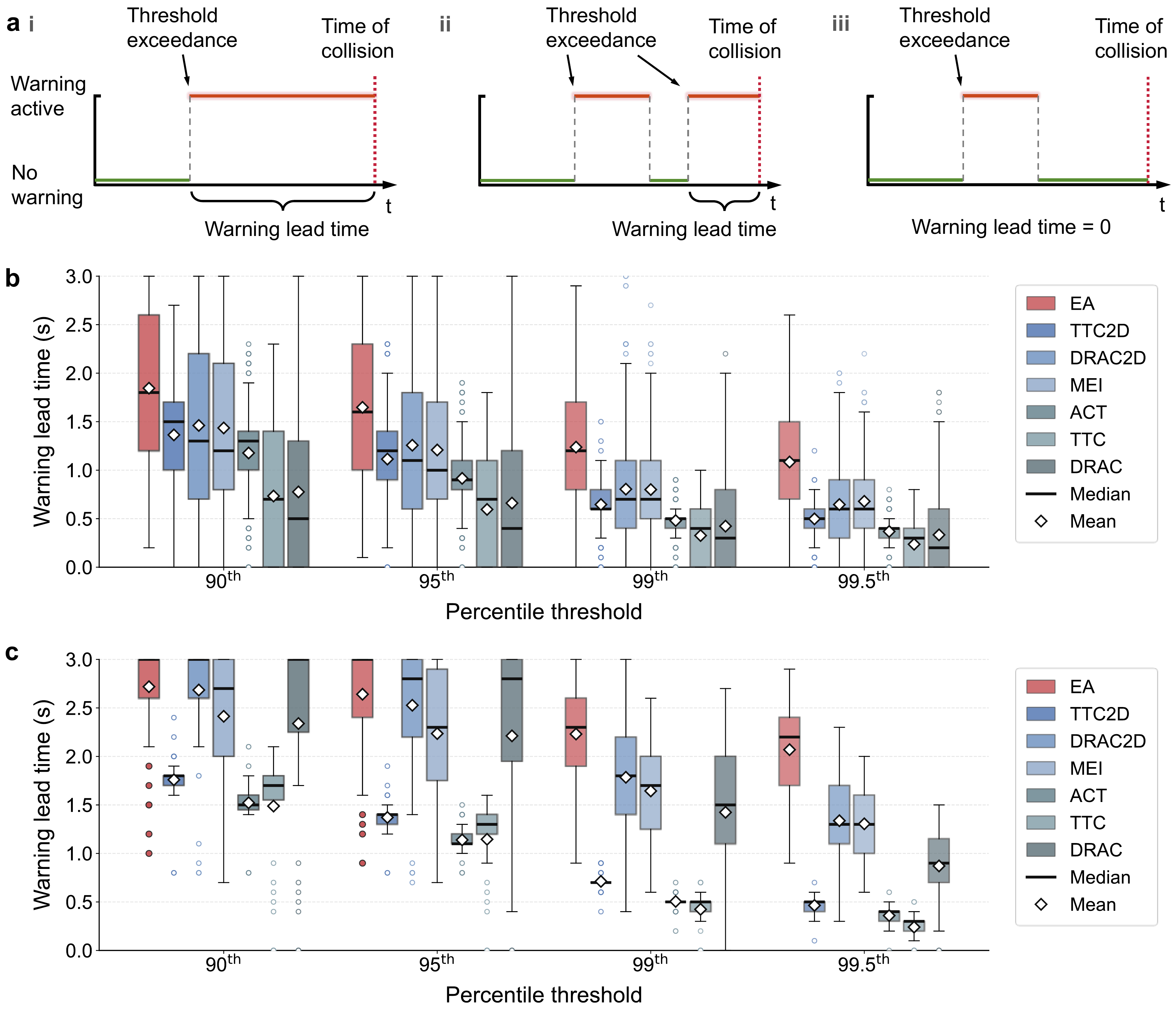}
    \caption{
    \textbf{Early warning performance under percentile-aligned false-alarm constraints.}
    \textbf{a,} Definition of warning lead time (WLT) under the sustained warning criterion. i, If threshold exceedance persists to the last effective precollision frame, WLT is measured from the first such exceedance. ii, If the warning disappears and later reappears, only the final sustained exceedance is counted. iii, If no sustained warning persists to the end, WLT is set to zero.
    \textbf{b,} Distribution of warning lead times across all 658 crash cases for four percentile-aligned thresholds (90th, 95th, 99th and 99.5th). EA consistently provides the earliest sustained warning, with a larger advantage at stricter thresholds.
    \textbf{c,} Warning lead times for the vehicle--pedestrian subset. EA shows a particularly strong advantage in these interactions, where TTC-based methods are more readily triggered by benign but close pedestrian approaches.
}
    \label{fig:Results_2_2}
\end{figure}

To ensure a rigorous and reproducible evaluation, we define the warning lead time (WLT) under the sustained warning criterion illustrated in Fig.~\ref{fig:Results_2_2}a and formally specified in the \hyperref[sec:methods_wlt]{Methods} section. A warning is counted as valid only if, once activated, it remains active until the last effective precollision frame. If a method crosses the threshold and then drops below it before collision, that earlier activation is discarded; only the final threshold exceedance that persists to the end is counted. If no such sustained activation exists, WLT is set to zero. This definition avoids assigning credit to unstable threshold crossings that would be operationally unreliable in practice.

To compare risk quantification methods under equal false-alarm constraints, thresholds are calibrated from the eventwise maximum risk value of each noncrash potential conflict case. Specifically, for each method, the 90th, 95th, 99th, and 99.5th percentiles of the noncrash event distribution are used as warning thresholds, hereafter denoted as P90, P95, P99, and P99.5, corresponding to event-level false-positive rates of approximately 10\%, 5\%, 1\%, and 0.5\%, respectively. For the four naturalistic datasets with available cumulative road-user exposure (inD, SIND, highD, and exiD), the translation of these percentile thresholds into expected alert intervals is provided in Supplementary Section~2b. In the urban data, the overall expected alert interval ranges from 3.8~min at P90 to 76.7~min at P99.5; in the highway data, it ranges from 11.0 to 219.7~min. These operating points therefore span a practically meaningful range from moderately frequent to highly conservative alerts.

\begin{table}[H]
\centering
\setlength{\tabcolsep}{4.5pt}
\renewcommand{\arraystretch}{1.15}

\caption{\textbf{Median warning lead times and EA's advantage over the baselines at percentile-aligned risk thresholds.} \textit{Median} (s) denotes the median warning lead time under the sustained warning criterion, with more negative values indicating earlier warning. $\Delta t$ (s) denotes Median$_{\rm other}$ $-$ Median$_{\rm EA}$, with positive values indicating an earlier warning by EA. \textit{Impv.} (\%) denotes the corresponding relative improvement.}
\label{tab:lead_time_stats_integrated}
\resizebox{\columnwidth}{!}{
\begin{tabular}{l*{12}{r}}
\toprule
\multirow{2}{*}{\textbf{Method}} 
& \multicolumn{12}{c}{\textbf{Risk percentile}} \\
\cmidrule(lr){2-13}
& \multicolumn{3}{c}{90th} 
& \multicolumn{3}{c}{95th} 
& \multicolumn{3}{c}{99th} 
& \multicolumn{3}{c}{99.5th} \\
\cmidrule(lr){2-4}\cmidrule(lr){5-7}\cmidrule(lr){8-10}\cmidrule(lr){11-13}
& Median (s) & $\Delta t$ (s) & Impv. (\%) 
& Median (s) & $\Delta t$ (s) & Impv. (\%) 
& Median (s) & $\Delta t$ (s) & Impv. (\%)
& Median (s) & $\Delta t$ (s) & Impv. (\%) \\
\midrule
EA   
& \textbf{\underline{-1.80}} & --- & --- 
& \textbf{\underline{-1.60}} & --- & --- 
& \textbf{\underline{-1.20}} & --- & --- 
& \textbf{\underline{-1.10}} & --- & --- \\
ACT   
& -1.30 & +0.50 & +38 
& -0.90 & +0.70 & +78 
& -0.50 & +0.70 & +140 
& -0.40 & +0.70 & +175 \\
TTC2D 
& \textbf{-1.50} & +0.30 & +20 
& \textbf{-1.20} & +0.40 & +33 
& -0.60 & +0.60 & +100 
& -0.50 & +0.60 & +120 \\
TTC   
& -0.70 & +1.10 & +157 
& -0.70 & +0.90 & +129 
& -0.40 & +0.80 & +200 
& -0.30 & +0.80 & +267 \\
DRAC  
& -0.50 & +1.30 & +260 
& -0.40 & +1.20 & +300 
& -0.30 & +0.90 & +300 
& -0.20 & +0.90 & +450 \\
DRAC2D
& -1.30 & +0.50 & +38 
& -1.10 & +0.50 & +45 
& \textbf{-0.70} & +0.50 & +71 
& \textbf{-0.60} & +0.50 & +83 \\
MEI   
& -1.20 & +0.60 & +50 
& -1.00 & +0.60 & +60 
& \textbf{-0.70} & +0.50 & +71 
& \textbf{-0.60} & +0.50 & +83 \\
\bottomrule
\end{tabular}
}
\end{table}

Under these percentile-aligned thresholds, EA consistently provides the earliest sustained warning among all compared methods. As shown in Fig.~\ref{fig:Results_2_2}b and summarized in Table~\ref{tab:lead_time_stats_integrated}, the median warning lead times of EA are $-1.80$, $-1.60$, $-1.20$, and $-1.10$~s at the 90th, 95th, 99th, and 99.5th percentiles, respectively, where more negative values indicate earlier warning. Compared with TTC2D, EA advances the median warning by 0.30~s at P90, 0.40~s at P95, and 0.60~s at both P99 and P99.5, corresponding to relative improvements of 20\%, 33\%, 100\%, and 120\%, respectively. The same pattern holds against other strong baselines: compared with ACT, EA warns 0.50--0.70~s earlier across all four thresholds, and compared with DRAC2D and MEI, the gain remains at 0.50--0.60~s from P95 onward. The gap widens monotonically as the false-alarm constraints become more stringent, indicating that EA preserves superior timeliness at low false-alarm rates, where warning reliability matters most.

Pedestrians are widely regarded as the most behaviourally uncertain road users. We therefore conduct an additional set of experiments focusing on vehicle--pedestrian interactions and crashes. As shown in Fig.~\ref{fig:Results_2_2}c, EA maintains markedly earlier warnings than all existing baselines across all four percentile thresholds, with an especially pronounced advantage over TTC-based methods. This pattern is consistent with the high variability of pedestrian behaviour: near approaches are common, but many remain dynamically easy to resolve. Under such conditions, TTC-based methods frequently yield small TTC values even in benign close-proximity encounters, making percentile-aligned TTC thresholds conservative and limiting their ability to distinguish truly dangerous precrash states at an early stage. As a result, high-risk events are often identified only shortly before impact, whereas EA remains more sensitive to genuinely safety-critical interactions.

Overall, under equal false-alarm constraints, EA consistently achieves the earliest warning, and its advantage increases as the warning threshold becomes more stringent. These results provide further evidence for EA: beyond stronger statistical separability, it also delivers superior practical warning timeliness, especially in the stringent operating regimes most relevant to safety-critical deployment. This pattern is consistent with EA's intervention-based formulation, which better distinguishes interactions that are merely close from those that are genuinely difficult to resolve.
\subsection{Retained information about eventual crash outcomes}
\label{subsec:ea_information_retention}

We next quantify how much information about the eventual crash-versus-noncrash outcome is retained in each method's instantaneous risk value, measured as the reduction in outcome uncertainty.

We formulate this analysis as a matched comparison on a common preevent time axis. The crash set consists of 658 real crash cases, each providing trajectories from 3.0\,s to 0.1\,s before impact. The noncrash set consists of potential-conflict episodes drawn from the large-scale dataset introduced above. Of the 44,180 extracted noncrash episodes, 40,417 contain a complete aligned window from $-3.0$\,s to $-0.1$\,s and are therefore included. Real crash cases are aligned by the true impact time ($t = 0$), whereas noncrash episodes are aligned by the time of minimum bounding-box distance between the two road users. This provides a method-independent anchor for noncrash episodes, avoiding any dependence on a particular risk quantification method for temporal alignment, and enables all episodes to be aligned on a unified preevent time axis. All episodes are then sampled on the same lead-time grid, $\tau \in \{-3.0, -2.9, \ldots, -0.1\}\,\mathrm{s}$, using nearest-frame matching. This design isolates differences in how each method maps the same interaction state to a risk value. Full implementation details are provided in the \hyperref[subsec:episode_construction_methods]{Methods} section.

\begin{figure}[H]
    \centering
    \includegraphics[width=0.95\textwidth]{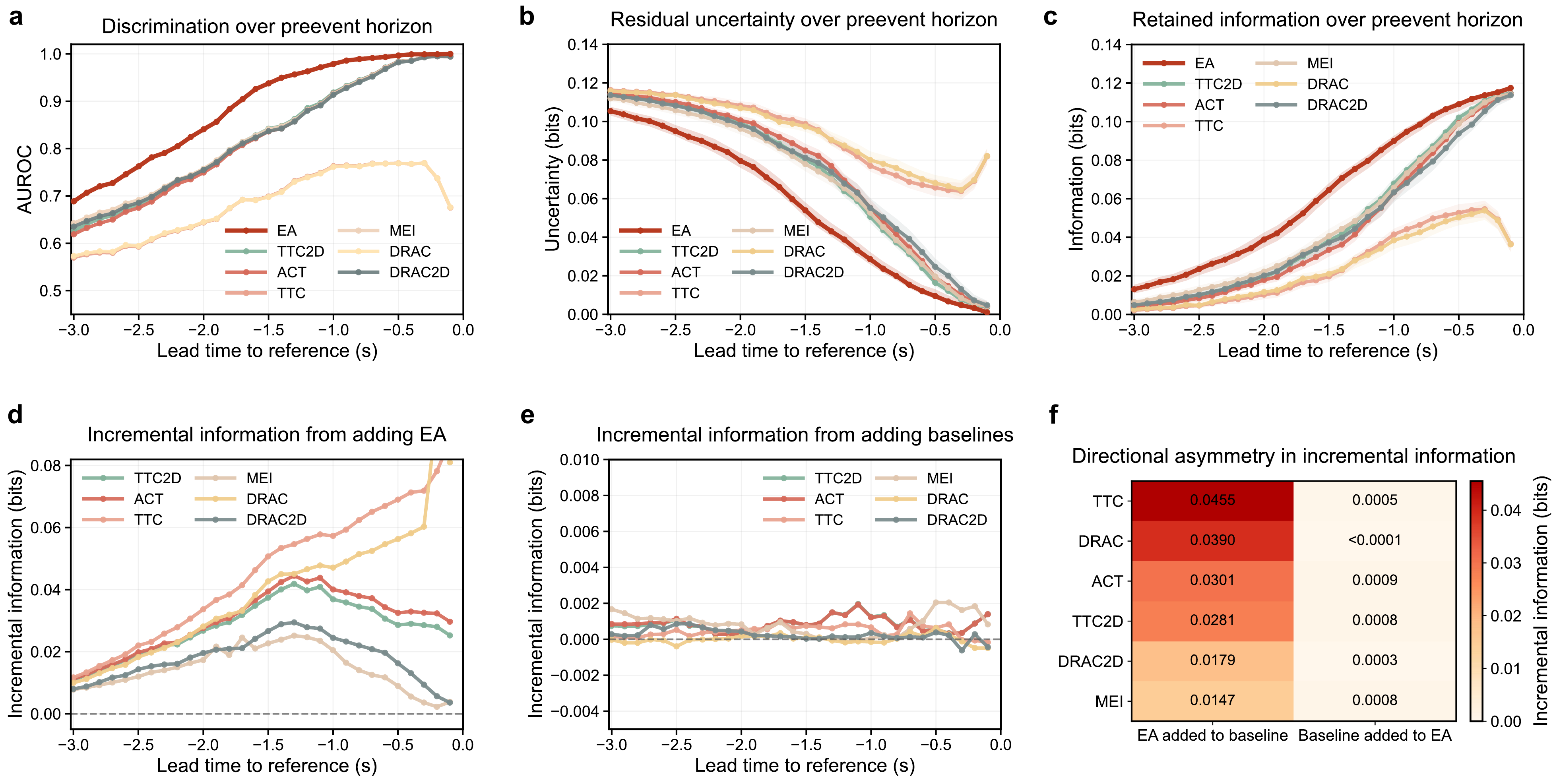}
    \caption{
    \textbf{EA exhibits the best standalone discrimination, the lowest residual uncertainty, the greatest retained information, and a pronounced directional asymmetry in incremental information relative to existing baselines.}
    All methods are evaluated on identical interaction states at each lead time. Real crash cases are aligned by the true impact time ($t = 0$), whereas noncrash potential-conflict episodes are aligned by the time of minimum bounding-box distance, serving as a method-independent reference. The analysis includes 658 crash cases and 40,417 noncrash episodes with complete aligned trajectories from $-3.0$\,s to $-0.1$\,s. Shaded bands in panels b and c denote case-level bootstrap 95\% confidence intervals.
    \textbf{a,} Standalone discrimination of eventual crash outcomes across the preevent horizon, measured by AUROC.
    \textbf{b,} Estimated residual uncertainty given each risk value, quantified by held-out predictive cross-entropy.
    \textbf{c,} Cross-entropy-based retained information in bits.
    \textbf{d,} Incremental information gained by adding EA to each baseline method.
    \textbf{e,} Incremental information gained by adding each baseline method on top of EA.
    \textbf{f,} Directional asymmetry in incremental information shown in panels d and e.}
    \label{fig:binary_information_main}
\end{figure}

For each lead time $\tau$, let $X_{\tau}$ denote the interaction state and let $M_j(\tau)=f_j(X_{\tau})$ denote the risk value returned by method $j$. We define retained information as the reduction in uncertainty relative to the baseline entropy:
\begin{equation}
\widehat{\mathcal{I}}^{\,\mathrm{ret}}_j(\tau)
=
H_{\tau}(Y)-\widehat{H}^{\,\mathrm{pred}}_{\tau}(Y\mid M_j),
\end{equation}
where $Y\in\{0,1\}$ denotes the binary eventual crash outcome and $H_{\tau}(Y)$ is the baseline entropy at lead time $\tau$. The term $\widehat{H}^{\,\mathrm{pred}}_{\tau}(Y\mid M_j)$ is the held-out predictive cross-entropy, interpreted here as the residual uncertainty after observing the method's risk value (see \hyperref[subsec:methods_information_retention]{Methods}). Because the matched dataset is highly imbalanced (658 crashes versus 40,417 noncrash episodes), the baseline entropy $H_{\tau}(Y)$ is inherently low, at approximately $0.1185$\,bits. This quantity defines the theoretical information ceiling imposed by the class imbalance; absolute retained-information values must therefore be interpreted relative to this ceiling. We report both residual uncertainty and retained information in bits. Representative lead times are summarised in Table~\ref{tab:retained_information_by_time} and Table~\ref{tab:ea_advantage_by_time}, directional asymmetry is summarised in Table~\ref{tab:directional_asymmetry}, and the full results at 0.1\,s resolution are provided in Supplementary Tables~4 and 5.

As shown in Fig.~\ref{fig:binary_information_main}a, EA achieves the best standalone discrimination across the full preevent horizon. Averaged over the 3\,s window, EA achieves a mean AUROC of 0.889, remaining above all recent baselines, including MEI (0.829), and substantially exceeding the classical baselines (TTC and DRAC, both 0.684). At $-3.0$\,s, EA already reaches an AUROC of approximately 0.69, compared with 0.62--0.64 for the recent baselines (ACT, MEI, TTC2D, and DRAC2D) and 0.57--0.58 for the classical baselines. EA therefore separates crash from noncrash interactions substantially earlier in the approach phase.

\begin{table}[H]
\caption{\textbf{Retained information at representative lead times.} For each method, \textit{Info.} denotes retained information in bits, and \textit{Ratio} denotes the percentage of the task ceiling retained. The best and second-best values per row are marked by bold underlining and boldface, respectively. The last row reports averages across all sampled lead times.}
\label{tab:retained_information_by_time}
\centering
\fontsize{8.5}{9}\selectfont
\renewcommand{\arraystretch}{1.20}
\setlength{\tabcolsep}{5pt}
\begin{tabular}{@{} l *{14}{r} @{}}
\toprule
\multirow{2}{*}{\textbf{Lead time}}
& \multicolumn{2}{c}{\textbf{EA}}
& \multicolumn{2}{c}{\textbf{ACT}}
& \multicolumn{2}{c}{\textbf{TTC2D}}
& \multicolumn{2}{c}{\textbf{TTC}}
& \multicolumn{2}{c}{\textbf{DRAC}}
& \multicolumn{2}{c}{\textbf{DRAC2D}}
& \multicolumn{2}{c}{\textbf{MEI}} \\
\cmidrule(lr){2-3}\cmidrule(lr){4-5}\cmidrule(lr){6-7}\cmidrule(lr){8-9}
\cmidrule(lr){10-11}\cmidrule(lr){12-13}\cmidrule(lr){14-15}
& \shortstack{Info.}
& \shortstack{Ratio\\(\%)}
& \shortstack{Info.}
& \shortstack{Ratio\\(\%)}
& \shortstack{Info.}
& \shortstack{Ratio\\(\%)}
& \shortstack{Info.}
& \shortstack{Ratio\\(\%)}
& \shortstack{Info.}
& \shortstack{Ratio\\(\%)}
& \shortstack{Info.}
& \shortstack{Ratio\\(\%)}
& \shortstack{Info.}
& \shortstack{Ratio\\(\%)} \\
\midrule
-3.0 s & \textbf{\underline{0.013}} & \textbf{\underline{11.0}} & 0.004 & 3.3 & 0.004 & 3.7 & 0.002 & 1.8 & 0.003 & 2.2 & 0.005 & 4.0 & \textbf{0.007} & \textbf{5.5} \\
-2.5 s & \textbf{\underline{0.024}} & \textbf{\underline{19.9}} & 0.008 & 7.1 & 0.010 & 8.2 & 0.005 & 3.9 & 0.005 & 4.0 & 0.010 & 8.6 & \textbf{0.013} & \textbf{10.7} \\
-2.0 s & \textbf{\underline{0.039}} & \textbf{\underline{32.8}} & 0.018 & 15.1 & 0.019 & 16.3 & 0.010 & 8.7 & 0.012 & 9.8 & 0.020 & 17.1 & \textbf{0.022} & \textbf{18.8} \\
-1.5 s & \textbf{\underline{0.065}} & \textbf{\underline{54.5}} & 0.034 & 28.3 & 0.038 & 32.3 & 0.020 & 16.7 & 0.021 & 17.8 & 0.037 & 31.4 & \textbf{0.041} & \textbf{34.2} \\
-1.0 s & \textbf{\underline{0.090}} & \textbf{\underline{75.9}} & 0.063 & 53.4 & \textbf{0.068} & \textbf{57.3} & 0.042 & 35.0 & 0.038 & 32.4 & 0.063 & 53.4 & 0.066 & 55.5 \\
-0.5 s & \textbf{\underline{0.109}} & \textbf{\underline{92.0}} & 0.099 & 83.7 & \textbf{0.102} & \textbf{86.2} & 0.053 & 44.5 & 0.050 & 42.5 & 0.094 & 79.2 & 0.099 & 83.4 \\
-0.1 s & \textbf{\underline{0.118}} & \textbf{\underline{99.2}} & 0.114 & 96.5 & \textbf{0.115} & \textbf{96.9} & 0.037 & 30.9 & 0.036 & 30.7 & 0.114 & 96.0 & 0.114 & 96.4 \\
\midrule
\begin{tabular}{@{}l@{}}Average over full\\ preevent horizon\end{tabular} & \textbf{\underline{0.063}} & \textbf{\underline{53.5}} & 0.045 & 37.8 & 0.048 & 40.1 & 0.025 & 20.8 & 0.024 & 20.5 & 0.045 & 38.2 & \textbf{0.048} & \textbf{40.8} \\
\bottomrule
\end{tabular}
\end{table}

The AUROC results establish EA's superior standalone discrimination. We next turn to the information-retention analysis, which asks not only whether methods rank cases correctly, but how much outcome-relevant information remains in the risk value itself. As shown in Fig.~\ref{fig:binary_information_main}b,c, EA achieves the lowest residual uncertainty and thus retains the most information across the preevent horizon. At $-3.0$\,s, EA already retains 0.0131\,bits (11.0\% of the ceiling), roughly twice the value of MEI, about three times the values of ACT, TTC2D and DRAC2D, and four to six times the values of TTC and DRAC. Averaged over the full preevent horizon, EA remains the best-performing method, retaining 0.063\,bits (53.5\% of the ceiling), compared with 0.048\,bits for MEI and TTC2D and about 0.024--0.045\,bits for the other baselines.

As impact approaches, retained information increases for all methods, but the trajectories remain separated. EA's retained information rises steadily, reaching 99.2\% of the information ceiling at $-0.1$\,s. The recent baselines also move closer to the ceiling near impact (retaining 96.0--96.9\%), but remain below EA. In contrast, the classical baselines capture only about 30\% of the information ceiling even immediately before impact. EA approaching the information ceiling near impact implies that its risk value preserves nearly all the information extractable from the interaction state under this formulation. Under held-out evaluation, EA therefore does more than merely improve ranking; it provides a systematically less lossy summary of the interaction state.

We next test whether EA merely correlates with existing methods or instead contributes nonredundant information. Panels d and e in Fig.~\ref{fig:binary_information_main} show the time-resolved incremental gains in both directions, whereas Table~\ref{tab:directional_asymmetry} summarises the corresponding time-averaged asymmetry over the full preevent horizon. Adding EA to every baseline yields consistently positive gains, particularly against the classical methods, but remaining robust even for the recent baselines. In contrast, adding these baseline methods on top of EA yields only negligible average gains, on the order of 0 to 0.001\,bits.

\begin{table}[H]
\caption{\textbf{EA's advantage in retained information over the baselines at representative lead times.} For each baseline, $\Delta I$ and \textit{Impv.} denote EA's absolute advantage (in bits) and relative improvement, respectively. Positive values indicate that EA retains more information than the baseline. The last row reports averages across all sampled lead times.}
\label{tab:ea_advantage_by_time}
\centering
\fontsize{8.5}{9}\selectfont
\renewcommand{\arraystretch}{1.20}
\setlength{\tabcolsep}{5pt}
\begin{tabular}{@{} l *{12}{r} @{}}
\toprule
\multirow{2}{*}{\textbf{Lead time}}
& \multicolumn{2}{c}{\textbf{ACT}}
& \multicolumn{2}{c}{\textbf{TTC2D}}
& \multicolumn{2}{c}{\textbf{TTC}}
& \multicolumn{2}{c}{\textbf{DRAC}}
& \multicolumn{2}{c}{\textbf{DRAC2D}}
& \multicolumn{2}{c}{\textbf{MEI}} \\
\cmidrule(lr){2-3}\cmidrule(lr){4-5}\cmidrule(lr){6-7}\cmidrule(lr){8-9}\cmidrule(lr){10-11}\cmidrule(lr){12-13}
& \shortstack{$\Delta I$}
& \shortstack{Impv.\\(\%)}
& \shortstack{$\Delta I$}
& \shortstack{Impv.\\(\%)}
& \shortstack{$\Delta I$}
& \shortstack{Impv.\\(\%)}
& \shortstack{$\Delta I$}
& \shortstack{Impv.\\(\%)}
& \shortstack{$\Delta I$}
& \shortstack{Impv.\\(\%)}
& \shortstack{$\Delta I$}
& \shortstack{Impv.\\(\%)} \\
\midrule
-3.0 s & +0.009 & +233 & +0.009 & +200 & +0.011 & +501 & +0.011 & +397 & +0.008 & +173 & +0.007 & +102 \\
-2.5 s & +0.015 & +181 & +0.014 & +144 & +0.019 & +410 & +0.019 & +399 & +0.013 & +130 & +0.011 & +86 \\
-2.0 s & +0.021 & +117 & +0.020 & +102 & +0.029 & +278 & +0.027 & +235 & +0.019 & +92 & +0.017 & +75 \\
-1.5 s & +0.031 & +92 & +0.026 & +68 & +0.045 & +227 & +0.044 & +207 & +0.027 & +73 & +0.024 & +59 \\
-1.0 s & +0.027 & +42 & +0.022 & +32 & +0.049 & +117 & +0.052 & +134 & +0.027 & +42 & +0.024 & +37 \\
-0.5 s & +0.010 & +10 & +0.007 & +7 & +0.056 & +107 & +0.059 & +116 & +0.015 & +16 & +0.010 & +10 \\
-0.1 s & +0.003 & +3 & +0.003 & +2 & +0.081 & +221 & +0.081 & +223 & +0.004 & +3 & +0.003 & +3 \\
\midrule
\begin{tabular}{@{}l@{}}Average over full\\ preevent horizon\end{tabular} & +0.019 & \textbf{+96.1} & +0.016 & \textbf{+75.1} & +0.039 & \textbf{+241.4} & +0.039 & \textbf{+220.3} & +0.018 & \textbf{+75.4} & +0.015 & \textbf{+54.2} \\
\bottomrule
\end{tabular}
\end{table}

The time-resolved analysis shows that EA's advantage over the recent baselines arises predominantly in the early and intermediate preevent phases. At these lead times, crash-relevant structure is already present, but the interaction has not yet entered the geometrically imminent regime. Near impact, the absolute gaps between EA and the recent baselines narrow because all methods approach the low-entropy ceiling. In contrast, EA's absolute advantage over the classical baselines remains substantial throughout the entire horizon.

\begin{table}[H]
\centering
\caption{\textbf{Directional asymmetry in incremental information between EA and the baselines over the full preevent horizon.} $B$ denotes the baseline under comparison. Percentages are normalised by the information ceiling of 0.1185\,bits. The asymmetry ratio is defined as $(EA \rightarrow B)/(B \rightarrow EA)$.}
\label{tab:directional_asymmetry}
\renewcommand{\arraystretch}{1.15}
\footnotesize
\begin{tabular}{lcccccc}
\toprule
Baseline &
EA $\rightarrow$ B (bit) &
EA $\rightarrow$ B (\%) &
B $\rightarrow$ EA (bit) &
B $\rightarrow$ EA (\%) &
Asym. gap (bit) &
Asym. ratio \\
\midrule
ACT    & 0.0301 & \textbf{25.4} & 0.0009   & 0.7 & 0.0292 & 35.2 \\
TTC2D  & 0.0281 & \textbf{23.7} & 0.0008   & 0.7 & 0.0272 & 33.8 \\
TTC    & 0.0455 & \textbf{38.4} & 0.0005   & 0.4 & 0.0450 & 95.5 \\
DRAC   & 0.0390 & \textbf{32.9} & <0.0001  & 0.0 & 0.0390 & -- \\
DRAC2D & 0.0179 & \textbf{15.1} & 0.0003   & 0.2 & 0.0176 & 63.5 \\
MEI    & 0.0147 & \textbf{12.4} & 0.0008   & 0.7 & 0.0138 & 17.5 \\
\bottomrule
\end{tabular}
\end{table}

As detailed in Table~\ref{tab:directional_asymmetry}, adding EA to existing baselines yields normalised gains ranging from 12.4\% to 38.4\% of the information ceiling, while the reverse operations yield less than 1\%. This profound directional asymmetry suggests that EA captures fundamental kinematic or geometric features of the crash mechanics that are fundamentally unobservable by traditional baselines.

Taken together, these results demonstrate that EA consistently retains more crash-outcome information than the existing baselines when evaluated on identical interaction states. This advantage is already evident at the earliest sampled lead times, persists throughout the approach phase, and remains present even immediately before impact. Moreover, it is largely nonredundant with the information captured by existing methods.
\section{Discussion}

We present EA, which reliably and efficiently quantifies risk in driving interactions. EA represents a new paradigm for risk quantification. It measures the minimum instantaneous evasive cost required to restore safety. Notably, despite its concise theoretical assumptions, our method demonstrates consistent advantages over existing baselines. This suggests that the minimum-cost evasive direction may intrinsically capture essential aspects of the underlying driving principles of traffic behaviour, analogous to the intuitive human instinct of harm avoidance. Our statistical results indicate that, even in real-world environments constrained by traffic regulations and road geometry, this minimum-cost direction closely reflects the underlying mechanisms of interaction risk. Beyond its primary role as a safety benchmark, EA also offers a new perspective for studying autonomous decision-making and interactive traffic behaviour.

EA is limited by its modelling assumptions. First, we assume that the same relative acceleration has the same cost in all directions. For application-specific use, this assumption can be refined. Future models may incorporate more detailed cost differences, such as differences in manoeuvrability across traffic participants, directional asymmetries in individual motion capabilities, and constraints imposed by traffic rules and road geometry. Second, our current framework considers pairwise interactions only. This is because current safety benchmarks are largely pairwise. In multi-agent traffic scenarios, the constraints become more complex. Future work should therefore extend EA to multi-vehicle interaction settings. Moreover, this study focuses on the probability of potential collisions, without considering collision severity.

EA has important practical value. It can serve as a safety benchmark for full-stack autonomous driving tasks related to driving safety. First, it can guide the generation of safer driving behaviours. Second, it provides a more accurate and reliable benchmark for safety evaluation. Third, it facilitates the realisation of a safer and more efficient future transportation system.
We release the EA-P90 dataset\footnote{EA-P90 dataset: \url{https://drive.google.com/drive/folders/1JGE76fxJyQwvCQTU5soV8YLZySyFQBv9?usp=sharing}} to support downstream applications from training to testing. It consists of the top 10\% highest-risk scenarios in the large-scale potential-conflict dataset, ranked by the maximum EA value of each scenario; further details are provided in Supplementary Section 3d.
\phantomsection
\section{Methods}

\subsection{Definition of evasive acceleration}
\label{subsec:ea_definition_methods}

For two road users $A$ and $B$ at time $t$, let
\begin{equation}
S_{AB}(t):=\{S_A(t),S_B(t)\}
\label{eq:stateAB_methods}
\end{equation}
denote their joint interaction state, where $S_A(t)$ and $S_B(t)$ are the instantaneous states of the two road users. We treat the interacting pair as a joint system and define evasive acceleration (EA) in relative-motion space rather than as a control input assigned to either road user individually. We consider planar interactions and represent each road user as a rigid body with a two-dimensional occupied region. Unless otherwise specified, this occupied region is modelled as an oriented bounding box (OBB).

We define EA over a future interval of interest $[t,t+T]$, with $T>0$. Over this interval, the future motions of the two road users are extrapolated from the current state under a specified motion model and without any additional interaction-induced evasive response, yielding extrapolated trajectories $\mathbf p_A^{0}(t+s)$ and $\mathbf p_B^{0}(t+s)$ for $s\in[0,T]$. The corresponding extrapolated relative trajectory is
\begin{equation}
\mathbf r^{0}(t+s):=\mathbf p_A^{0}(t+s)-\mathbf p_B^{0}(t+s),
\qquad s\in[0,T].
\label{eq:r0_methods}
\end{equation}

Let $\mathcal C\subset\mathbb{R}^{2}$ denote the collision set in relative-position space, that is, the set of relative positions for which the occupied regions of the two road users overlap. We then introduce a hypothetical constant relative acceleration
\begin{equation}
\mathbf a=
\begin{bmatrix}
a_x\\
a_y
\end{bmatrix}
\in\mathbb{R}^{2},
\label{eq:a_methods}
\end{equation}
defined directly in relative-motion space. Under this intervention, the relative trajectory becomes
\begin{equation}
\mathbf r_{\mathbf a}(t+s)
=
\mathbf r^{0}(t+s)+\frac{1}{2}\mathbf a s^2,
\qquad s\in[0,T].
\label{eq:r_control_methods}
\end{equation}

EA is defined as the minimum Euclidean norm of a constant relative acceleration that keeps the future relative trajectory outside the collision set throughout the interval of interest:
\begin{equation}
\mathrm{EA}\!\left(S_{AB}(t)\right)
=
\min_{\mathbf a\in\mathbb{R}^{2}}
\|\mathbf a\|_2
\quad
\text{s.t.}
\quad
\mathbf r^{0}(t+s)+\frac{1}{2}\mathbf a s^2\notin\mathcal C,
\ \forall s\in[0,T].
\label{eq:EA_opt_methods}
\end{equation}

Thus, EA is a coupled quantity defined in relative-motion space and reflects the minimum evasive effort required to restore safety from the current interaction state. Its operational formulation and computation depend on the motion model used for extrapolation and are developed in the next subsection.

\subsection{Operational formulation and tractable computation of EA}
\label{subsec:ea_framework_methods}

As defined in Eq.~\eqref{eq:EA_opt_methods}, EA is a continuous-time rigid-body collision-avoidance optimisation problem in relative-motion space, which is generally non-trivial to solve efficiently. Because the extrapolated relative trajectory depends on the assumed motion model, EA admits different operational formulations under different short-horizon motion hypotheses. In this study, we develop a framework that combines analytical reduction with structured numerical evaluation, thereby preserving this definition while making EA computation tractable for large-scale evaluation. We consider two commonly used motion models: constant velocity (CV) and constant turn-rate and velocity (CTRV).

Under the CV model, the future state of a road user satisfies
\begin{equation}
\mathbf{x}(t+s)=\mathbf{x}(t)+\mathbf{v}(t)s,
\qquad s\in[0,T],
\label{eq:cv_model_methods}
\end{equation}
where $\mathbf{x}$ and $\mathbf{v}$ denote position and velocity, respectively. In naturalistic traffic interactions, especially in merging and turning scenarios, extrapolated trajectories often exhibit non-negligible curvature. Under CV extrapolation, lateral acceleration induced by the curvature of the extrapolated trajectory may therefore be attributed to evasive demand, leading to overestimation of the truly additional acceleration required for collision avoidance. To reduce this mismatch, we additionally consider EA under the CTRV model~\cite{polychronopoulos2007sensor,schubert2008comparison,schubert2011empirical,baek2020vehicle}. Under the CTRV assumption, motion evolves according to
\begin{equation}
\dot{x}(t)=v(t)\cos\theta(t), \quad
\dot{y}(t)=v(t)\sin\theta(t), \quad
\dot{\theta}(t)=\omega,
\label{eq:ctrv_model_methods}
\end{equation}
where $\omega$ denotes the constant yaw rate, and speed is assumed constant over the interval of interest.

Only the CV--CV case admits the analytical reduction developed in this study. Formulations involving CTRV motion are evaluated numerically using directional discretisation and discrete-time collision checking.

Under the CV--CV motion combination, we introduce a radial--tangential (R--T) coordinate system as a local frame centred at road user $B$, with its radial axis aligned with the instantaneous relative velocity of $A$ with respect to $B$ and its tangential axis given by the orthogonal direction. Let $\mathbf{v}_{\mathrm{rel}}:=\mathbf{v}_A-\mathbf{v}_B$ and $v_R:=\|\mathbf{v}_{\mathrm{rel}}\|_2$, where $v_R>0$ denotes the relative speed. In this frame, the constant relative acceleration is written as $\mathbf a=(a_R,a_T)$, and $\mathrm{EA}_{\mathrm{CV}}$ is evaluated as the minimum Euclidean norm of a constant relative acceleration that keeps the future relative motion collision-free under CV extrapolation.

For the upward-evasion mode, the CV-based EA problem can be written as
\begin{equation}
EA_U
=
\min_{\mathbf a \in \mathrm{SR}(\mathcal B_U)}
\|\mathbf a\|_2,
\qquad
\mathrm{SR}(\mathcal B_U)
=
\bigcap_{B_i\in\mathcal B_U}\mathrm{SR}(B_i),
\label{eq:EAU_methods}
\end{equation}
where $\mathcal B_U$ denotes the set of Step Barriers induced by the sampled danger-side boundary points, and $\mathrm{SR}(B_i)$ denotes the collision-avoiding acceleration set associated with a single Step Barrier $B_i=(d_{R,i},d_{T,i})$. The downward-evasion mode is defined symmetrically, and the global CV-based EA is
\begin{equation}
EA_{\mathrm{CV}}=\min\{EA_U,EA_D\}.
\label{eq:EA_global_methods}
\end{equation}

The key simplification is that the original continuous-time rigid-body collision-avoidance problem can be reduced to analytically tractable single-barrier subproblems. For each Step Barrier $B_i=(d_{R,i},d_{T,i})$ in the upward-evasion mode, the collision-free set in $(a_R,a_T)$ space is characterized by a lower boundary on $a_R$ as a function of $a_T$:
\begin{equation}
BC_i(a_T)=
\begin{cases}
a^\star_{R,i}, & 0\le a_T<a^\star_{T,i},\\[0.8ex]
v_R\sqrt{\dfrac{2a_T}{d_{T,i}}}-\dfrac{d_{R,i}}{d_{T,i}}\,a_T,
& a_T\ge a^\star_{T,i},
\end{cases}
\label{eq:BCi_methods}
\end{equation}
with
\begin{equation}
a^\star_{R,i}=\frac{v_R^2}{2d_{R,i}},
\qquad
a^\star_{T,i}=\frac{d_{T,i}v_R^2}{2d_{R,i}^2}.
\label{eq:a_star_methods}
\end{equation}
Thus, each single-barrier constraint becomes a lower bound on $a_R$ as a function of $a_T$: for small $a_T$, safety is achieved by sufficient radial deceleration alone, whereas for larger $a_T$, safety is achieved through coordinated radial--tangential avoidance.

Since the multi-barrier collision-avoiding acceleration set is the intersection of the corresponding single-barrier sets, the boundary of the global collision-avoiding acceleration set for the upward mode is given by
\begin{equation}
BC_U(a_T):=\max_{B_i\in\mathcal B_U}BC_i(a_T),
\qquad a_T\ge 0.
\label{eq:BCU_methods}
\end{equation}
The optimisation in Eq.~\eqref{eq:EAU_methods} is therefore reduced to finding the minimum-norm point on or above the upper envelope $BC_U(a_T)$.

After removing dominated Step Barriers, the global optimum can be found by evaluating a finite candidate set consisting of the radial optimum of the active barrier family, single-barrier tangential optima that remain feasible under all retained barriers, and pairwise intersections of active boundary curves lying on the global envelope $BC_U$. Letting $\mathcal C_U$ denote the union of these candidate points, the optimal upward-mode solution is obtained as
\begin{equation}
\mathbf a_U^{\mathrm{opt}}
=
\arg\min_{\mathbf a\in\mathcal C_U}\|\mathbf a\|_2,
\qquad
EA_U=\|\mathbf a_U^{\mathrm{opt}}\|_2.
\label{eq:finite_candidate_methods}
\end{equation}
The same construction applies to the downward-evasion mode, yielding $\mathbf a_D^{\mathrm{opt}}$ and $EA_D$, after which the final CV-based EA value follows from Eq.~\eqref{eq:EA_global_methods}. This reduction transforms the original continuous-time, multi-constraint collision-avoidance problem into a finite candidate evaluation problem in acceleration space, thereby making the CV-based EA computation both exact under the CV assumption and highly efficient. The full derivation, including Step Barrier construction, dominance conditions, and closed-form boundary-intersection formulas, is provided in Supplementary Section~1a.

To extend EA beyond straight-line extrapolation, we consider motion formulations in which one or both interacting road users follow CTRV dynamics over the interval of interest. This includes the $(\mathrm{CTRV},\mathrm{CTRV})$ formulation as well as the mixed cases $(\mathrm{CV},\mathrm{CTRV})$ and $(\mathrm{CTRV},\mathrm{CV})$, with the CV road user treated as the special case of zero yaw rate.

Consistent with the EA definition above, we introduce a constant relative acceleration $\mathbf a=(a_x,a_y)\in\mathbb{R}^2$ as a system-level intervention in relative-motion space. In implementation, this is realised as a quadratic translation of one predicted centre trajectory while retaining the baseline orientation evolution under the chosen motion model. If $\mathbf p^{(A)}(t+s)$ denotes the baseline predicted centre of road user $A$, then under $\mathbf a$ the translated centre is
\begin{equation}
\tilde{\mathbf p}^{(A)}(t+s;\mathbf a)
=
\mathbf p^{(A)}(t+s)+\frac{1}{2}\mathbf a\,s^2.
\label{eq:ctrv_shift_main}
\end{equation}
The corresponding collision-avoiding acceleration set is defined by requiring the translated occupancy region of road user $A$ to remain disjoint from that of road user $B$ throughout the interval of interest. If $\mathbf 0$ already satisfies this condition, the corresponding EA value is zero.

Under CTRV-based extrapolation, the time-varying curvature and orientation of the occupied regions generally preclude an envelope-form solution. We therefore evaluate EA numerically by directional discretisation in acceleration space. Specifically, the relative acceleration is restricted to rays of the form
\begin{equation}
\mathbf a = m\,\mathbf u(\phi),
\qquad
m\in[0,a_{\max}],
\label{eq:ctrv_ray_main}
\end{equation}
and the problem is reduced to determining, for each direction $\phi$, the smallest magnitude $m$ for which $m\,\mathbf u(\phi)$ is collision-free over the interval of interest.

For a fixed direction, collision is assessed on a discretized prediction grid using an exact OBB overlap test based on the Separating Axis Theorem. At each sampled prediction time, the overlap constraints induce an interval of acceleration magnitudes along that direction that still lead to collision. The union of these intervals over the interval of interest therefore defines the directional collision set. The directional minimum is then given by the smallest admissible magnitude lying strictly outside the connected collision component that contains the origin. Repeating this evaluation over a finite set of candidate directions yields the CTRV-based EA as
\begin{equation}
\mathrm{EA}_{\mathrm{CTRV}}
=
\min_{\phi\in\Phi} m(\phi),
\label{eq:ctrv_ea_main}
\end{equation}
where $\Phi$ denotes the discretised directional set and $m(\phi)$ the corresponding directional minimum. In practice, we adopt a coarse-to-fine directional evaluation, first sweeping the full angular domain coarsely and then refining the search locally around the best coarse direction. For each fixed direction, the minimum feasible magnitude is obtained by interval evaluation rather than iterative one-dimensional optimisation. If no admissible magnitude exists for any candidate direction within the prescribed acceleration bound, the corresponding EA value is treated as undefined under that bound. Detailed procedures for this numerical evaluation are provided in Supplementary Section~1b.

Because both CV and CTRV are simplified short-horizon motion hypotheses, either assumption alone may introduce systematic bias in some interaction patterns. We therefore adopt a simple model-averaging strategy so that the resulting EA value is not tied to a single extrapolation hypothesis. Allowing each interacting road user to follow either CV or CTRV yields four motion combinations:
\begin{equation}
\mathcal{M}=
\{(\mathrm{CV},\mathrm{CV}),
(\mathrm{CV},\mathrm{CTRV}),
(\mathrm{CTRV},\mathrm{CV}),
(\mathrm{CTRV},\mathrm{CTRV})\}.
\label{eq:model_set_methods}
\end{equation}
For each $m\in\mathcal{M}$, we compute the corresponding $\mathrm{EA}_m(t)$ and define the final EA as the arithmetic mean:
\begin{equation}
\mathrm{EA}(t)=
\frac{1}{|\mathcal{M}|}\sum_{m\in\mathcal{M}}\mathrm{EA}_m(t).
\label{eq:model_average_methods}
\end{equation}
This equal-weight aggregation provides a model-robust operationalisation of EA across heterogeneous traffic scenarios by reducing sensitivity to any single short-horizon motion assumption.

To evaluate computational efficiency, we measure runtime in Python on the 4,418 scenarios in the EA-P90 dataset. For the interval-of-interest setting $T=7.0\,\mathrm{s}$, the final EA defined by the four-model average in Eq.~\eqref{eq:model_average_methods} requires an average of $5.012$~ms per frame. These results indicate that the proposed framework is computationally lightweight. More importantly, they show that the structured combination of analytical reduction and controlled numerical evaluation makes the EA definition operational at scale while retaining fidelity to the underlying collision-avoidance formulation.

\subsection{Screening of potential conflicts}
\label{potential_conflicts}

Potential conflicts are screened using the following criteria:
\begin{enumerate}
    \item The two road users are observed in the same spatial region during an overlapping time interval.
    \item At least one frame satisfies any of the following criteria: $\mathrm{TTC} \le 5$~s, $\mathrm{ACT} \le 5$~s, or $\mathrm{TTC2D} \le 5$~s.
    \item At least one frame satisfies a bounding-box distance between the two road users of no more than $50$~m.
\end{enumerate}

Potential conflict screening commonly relies on time-based methods, particularly TTC and its variants. To retain potentially safety-relevant interactions, we intentionally adopt a conservative pre-screening strategy. Specifically, we use a temporal threshold of $5$~s, which is more inclusive than many commonly used TTC-based criteria, such as $1.5$~s~\cite{brow1994traffic,el2013safety,papadoulis2019evaluating}, $2.0$~s~\cite{sayed1999traffic,zhang2017safety,li2017evaluating}, and $3.0$~s~\cite{rahman2018longitudinal,rahman2019safety,wang2021review}; related studies have also suggested practical conflict-defining ranges on the order of $1$--$3$~s~\cite{rahman2018longitudinal,rahman2019safety,wang2021review}.

\subsection{Formal definition of warning lead time under the sustained-warning criterion}
\label{sec:methods_wlt}

To ensure consistent and reproducible evaluation of early-warning performance, we define warning lead time (WLT) under a sustained-warning criterion. Under this criterion, a warning is considered valid only if, once triggered, it remains active until the last valid precollision observation time. This definition excludes transient threshold exceedances. The three schematic cases implied by this criterion are illustrated in Fig.~\ref{fig:Results_2_2}a.

Let $t_c$ denote the collision time, and let $t_m \leq t_c$ denote the last valid precollision observation time, which in discretely sampled data is typically the final annotated frame before collision. For a given risk quantification method with risk trajectory $R(t)$ and warning threshold $\theta$, we define the warning indicator as
\begin{equation}
W(t)=\mathbf{1}\{R(t)\geq \theta\}.
\label{eq:warning_indicator_methods}
\end{equation}

A time $s \leq t_m$ is called a sustained-warning entry time if the warning remains active throughout the interval from $s$ to $t_m$. We therefore define the set of sustained-warning entry times as
\begin{equation}
\mathcal{A}(R,\theta)
=
\left\{
\, s \leq t_m
\;\middle|\;
W(\tau)=1,\ \forall \tau \in [s,t_m]
\right\}.
\label{eq:sustained_warning_set_methods}
\end{equation}

If $\mathcal{A}(R,\theta)\neq\emptyset$, the effective warning onset is defined as
\begin{equation}
t^{\ast}(R,\theta)=\inf \mathcal{A}(R,\theta).
\label{eq:warning_onset_methods}
\end{equation}

The warning lead time is then defined as
\begin{equation}
\mathrm{WLT}(R,\theta)
=
\begin{cases}
t_m-t^{\ast}(R,\theta), & \mathcal{A}(R,\theta)\neq\emptyset,\\[4pt]
0, & \mathcal{A}(R,\theta)=\emptyset.
\end{cases}
\label{eq:wlt_methods}
\end{equation}

Thus, only the final warning segment that persists until $t_m$ is credited; if no such sustained activation exists, WLT is set to zero. This logic corresponds directly to the three schematic cases shown in Fig.~\ref{fig:Results_2_2}a.

\subsection{Construction and temporal alignment of crash and noncrash episodes}
\label{subsec:episode_construction_methods}

The crash-outcome analyses are based on two episode sets: real crash episodes and noncrash potential conflict episodes. The crash set comprises 658 real collision cases. For each case, per-frame risk values are available from $3.0$\,s before impact to $0.1$\,s before impact. The noncrash set is drawn from the extracted potential conflict dataset. Of the 44,180 extracted episodes, 40,417 contain a complete preevent window covering the interval from $-3.0$\,s to $-0.1$\,s after alignment and are retained for analysis.

For crash episodes, the alignment anchor is the true impact time, defined as $t=0$. For noncrash potential conflict episodes, no true impact time exists, so we use the time of minimum bounding-box distance between the two interacting road users as a common geometric reference:
\begin{equation}
t_{\mathrm{min}}=
\arg\min_{t} d_{\mathrm{bbox}}(t),
\label{eq:bbox_anchor_methods}
\end{equation}
where $d_{\mathrm{bbox}}(t)$ is the instantaneous minimum distance between the occupied rigid-body bounding boxes. This anchor $t_{\mathrm{min}}$ is then shifted to $t=0$ to align with the collision timescale. Because it is defined purely geometrically, it is independent of all compared risk quantification methods and should be interpreted only as a common alignment reference, not as a claim that closest approach coincides with the peak latent-risk moment.

After alignment, all crash and noncrash episodes are sampled on the same lead-time grid relative to $t=0$,
\begin{equation}
\tau \in \{-3.0,-2.9,\ldots,-0.1\}\ \mathrm{s}.
\label{eq:leadtime_grid_methods}
\end{equation}
For each target lead time, the nearest available frame is selected within a tolerance of 0.03\,s. This procedure is applied at the case level, so that each included episode contributes at most one sampled frame to each lead-time slice.

\subsection{Evaluation of crash-outcome information retention}
\label{subsec:methods_information_retention}

This subsection describes the crash-versus-noncrash information-retention analysis reported in the Results section. We compare how much information about eventual crash outcome is retained in the risk values produced by different risk quantification methods.

At each sampled lead time $\tau$, each compared risk quantification method is evaluated on the same interaction snapshot. For each method $M_j$ and lead time $\tau$, we estimate the probability of eventual crash outcome using case-level out-of-fold prediction. All crash and noncrash episodes in the corresponding lead-time slice are partitioned by case into stratified folds. Within each fold, the scalar-to-probability mapping is fitted on the remaining folds and then used to generate predictions for the held-out cases. This procedure is repeated independently for each method and each lead-time slice, and all reported uncertainty quantities are computed from the pooled out-of-fold predictions.

For single-method analysis, the scalar-to-probability mapping is implemented using logistic regression with second-order polynomial terms applied to the raw risk value after fold-wise preprocessing. For incremental information analysis, we additionally fit a joint two-method mapping that combines EA with a baseline method using the same nested model family, including two linear terms, their quadratic terms, and their interaction term.

Let $Y\in\{0,1\}$ denote crash versus noncrash outcome. Within each lead-time slice $\tau$, let $H_{\tau}(Y)$ denote the binary entropy of the outcome distribution in that slice. The residual uncertainty after observing the value of method $M_j$ is estimated by the empirical held-out predictive cross-entropy,
\begin{equation}
\widehat{H}^{\,\mathrm{pred}}_{\tau}(Y\mid M_j)
=
-\frac{1}{n_{\tau}}
\sum_{i=1}^{n_{\tau}}
\Big[
y_i \log_2 \hat p_{ij,\tau}
+
(1-y_i)\log_2(1-\hat p_{ij,\tau})
\Big],
\label{eq:predictive_cross_entropy_methods}
\end{equation}
where $\hat p_{ij,\tau}$ is the out-of-fold calibrated crash probability for sample $i$ at lead time $\tau$. The corresponding information retained is then defined as
\begin{equation}
\widehat{\mathcal{I}}^{\,\mathrm{ret}}_j(\tau)
=
H_{\tau}(Y)-\widehat{H}^{\,\mathrm{pred}}_{\tau}(Y\mid M_j).
\label{eq:predictive_information_methods}
\end{equation}

To quantify nonredundant information, we compare the held-out predictive cross-entropy of a baseline method $M_k$, EA alone, and the joint mapping combining EA and $M_k$. The incremental information retained from adding EA to $M_k$ is defined as
\begin{equation}
\Delta \widehat{\mathcal{I}}^{\,\mathrm{ret}}_{\mathrm{EA}\mid k}(\tau)
=
\widehat{H}^{\,\mathrm{pred}}_{\tau}(Y\mid M_k)
-
\widehat{H}^{\,\mathrm{pred}}_{\tau}(Y\mid \mathrm{EA},M_k),
\label{eq:added_predictive_info_ea_given_k_methods}
\end{equation}
and the converse quantity, obtained by adding $M_k$ on top of EA, is defined analogously. A larger positive value indicates that the added method provides greater nonredundant information beyond the baseline mapping.

To quantify sampling uncertainty, we use case-level bootstrap resampling within the crash and noncrash groups and rerun the full information-retention analysis in each replicate. The shaded bands in Fig.~\ref{fig:binary_information_main}b,c show the resulting 95\% bootstrap confidence intervals.

\section*{Data availability}

The naturalistic driving datasets used in this study are highD, inD, exiD, SIND, and the Lateral Conflict Resolution Dataset (derived from Argoverse-2), which are available at
\url{https://www.highd-dataset.com},
\url{http://www.inD-dataset.com},
\url{https://levelxdata.com/exid-dataset/},
\url{https://github.com/SOTIF-AVLab/SinD}, and
\url{https://github.com/RomainLITUD/conflict_resolution_dataset}, respectively.

\section*{Code availability}


The source code for EA is available at \url{https://github.com/AutoChengh/evasive-acceleration}.

\bibliography{sample}

\section*{Acknowledgements}

This work was supported by the National Natural Science Foundation of China (Grant Nos.\ 52572497 and 52572477), the National Key R\&D Program of China (Grant No.\ 2023YFB2504400), and the Independent Research Project of the State Key Laboratory of Intelligent Green Vehicle and Mobility, Tsinghua University (Grant No.\ ZZ-PY-20250409). The authors also sincerely thank Dr.\ Yiru Jiao for her valuable guidance on the validation.

\section{Author contributions}
H.C. and S.Z. conceived and led the entire research project and proposed the fundamental concept of quantifying the minimum collision-avoidance cost as a measure of risk. H.C., Y.J., and W.Y. developed the algorithms, processed the dataset, organized the results, and wrote the paper. Z.L. and K.C. provided support for the preliminary validation of the work. R.Z. and J.B. contributed crash data. F.Z. provided support for the real-road experiments. H.H., H.Z., and J.W. offered important suggestions on algorithm development and experimental procedures. All authors contributed to manuscript revision and discussion of the results. S.Z. approved the submission and takes responsibility for the overall integrity of the paper.

\section{Competing interests}
The authors declare that a Chinese patent application is being filed for the analytical solution method for EA under the constant-velocity model described in the Methods section. The authors declare no other competing interests.

\end{document}